\documentclass{article} 
\usepackage{iclr2026_conference,times}
\usepackage{easyReview}

\usepackage{amsmath,amsfonts,bm}

\def\eqref#1{equation~\ref{#1}}

\def\1{\bm{1}}

\DeclareMathAlphabet{\mathsfit}{\encodingdefault}{\sfdefault}{m}{sl}
\SetMathAlphabet{\mathsfit}{bold}{\encodingdefault}{\sfdefault}{bx}{n}

\DeclareMathOperator*{\argmin}{arg\,min}

\usepackage{hyperref}
\usepackage{url}

\usepackage{amsfonts}       
\usepackage{nicefrac}       
\usepackage{microtype}      
\usepackage{xcolor}         

\usepackage{microtype}
\usepackage{graphicx}
\usepackage{subfigure}
\usepackage{booktabs} 

\usepackage{multirow,colortbl}
\definecolor{grey}{rgb}{0.9,0.9,0.9}
\usepackage{wrapfig}
\usepackage{tcolorbox}      
\usepackage{tikz}

\usepackage{amsmath}
\usepackage{amssymb}
\usepackage{mathtools}
\usepackage{amsthm}
\usepackage{upgreek}

\usepackage[capitalize]{cleveref}
\usepackage{subcaption}
\usepackage{subfigure}
\usepackage{csquotes}

\usepackage{xspace}
\usepackage{hypcap} 
\usepackage{bbm}
\usepackage{amsthm}

\usepackage{textcase}
\usepackage[normalem]{ulem}

\theoremstyle{plain}

\theoremstyle{definition}

\theoremstyle{remark}

\newtheorem*{proposition*}{Proposition}
\newtheorem*{theorem*}{Theorem}

\graphicspath{{figure/}{figures/}}
\newcommand{\md}{\text{d}}

\newcommand{\testdist}{p_{\text{te}}}
\newcommand{\textsfit}[1]{\textit{\textsf{#1}}}

\usepackage{xfrac}
\usepackage{listings}
\usepackage{xcolor}

\usepackage{algorithm}
\usepackage{algorithmicx,algpseudocode}

\usepackage{multicol}

\algblockdefx{MRepeat}{EndRepeat}{\textbf{Repeat}}{}
\algnotext{EndRepeat}

\definecolor{rebutcolor}{rgb}{0,0,0}

\definecolor{codegreen}{rgb}{0,0.6,0}
\definecolor{codegray}{rgb}{0.5,0.5,0.5}
\definecolor{codepurple}{rgb}{0.58,0,0.82}
\definecolor{backcolour}{rgb}{0.95,0.95,0.92}

\lstdefinestyle{mystyle}{
    backgroundcolor=\color{backcolour},   
    commentstyle=\color{codegreen},
    keywordstyle=\color{magenta},
    numberstyle=\tiny\color{codegray},
    stringstyle=\color{codepurple},
    basicstyle=\ttfamily\scriptsize,
    breakatwhitespace=false,         
    breaklines=true,                 
    captionpos=b,                    
    keepspaces=true,                 
    numbers=left,                    
    numbersep=5pt,                  
    showspaces=false,                
    showstringspaces=false,
    showtabs=false,                  
    tabsize=2,
}

\usepackage{float}

\usepackage{etoolbox}
\usepackage{pifont}
\usepackage{cancel}
\usepackage{placeins}
\usepackage{makecell}

\usepackage{tabularx,ragged2e}

\usepackage{bm} 

\setcitestyle{authoryear,square}

\usepackage{titletoc}
\usepackage[subfigure]{tocloft}
\usepackage{textcase}

\usepackage[dvipsnames,svgnames]{xcolor}
\usepackage{adjustbox}

\PassOptionsToPackage{textsize=tiny}{todonotes}
\definecolor{rev_red}{HTML}{D62728}
\definecolor{rev_blue}{HTML}{1F77B4}
\definecolor{rev_green}{HTML}{2CA02C}
\definecolor{rev_purple}{HTML}{9467BD}
\definecolor{rev_orange}{HTML}{FF7F0E} 

\reversemarginpar 
\let\oldfootnotemark\footnotemark
\renewcommand\footnotemark{%
  \begingroup
  \hypersetup{linkcolor=Maroon}
  \oldfootnotemark
  \endgroup
}

\hypersetup{
    colorlinks=true,
    linkcolor=teal, 
    citecolor=DarkBlue,   
    urlcolor=Maroon   
}

\title{Out-of-Support Generalisation via Weight-Space Sequence Modelling}
\makeatletter
\let\thetitle\@title
\let\theauthor\@author
\makeatother

\renewcommand{\cite}{\citep}

\renewcommand*{\bibfont}{\small}

\iclrfinalcopy

\author{Roussel Desmond Nzoyem \\
University of Bristol\\
Bristol, UK \\
\texttt{rd.nzoyemngueguin@bristol.ac.uk} \\
}

\newcolumntype{K}[1]{>{\centering\arraybackslash}p{#1}}

\renewcommand{\cite}{\citep}

\newcommand{\themethod}{\text{WeightCaster} }

\begin{document}

\maketitle

\begin{abstract}


As breakthroughs in deep learning transform key industries, models are increasingly required to extrapolate on datapoints found outside the range of the training set, a challenge we coin as out-of-support (OoS) generalisation. However, neural networks frequently exhibit catastrophic failure on OoS samples, yielding unrealistic but overconfident predictions. We address this challenge by reformulating the OoS generalisation problem as a sequence modelling task in the weight space, wherein the training set is partitioned into concentric shells corresponding to discrete sequential steps. Our \themethod framework yields plausible, interpretable, and uncertainty-aware predictions without necessitating explicit inductive biases, all the while maintaining high computational efficiency. Emprical validation on a synthetic cosine dataset and real-world air quality sensor readings demonstrates performance competitive or superior to the state of the art. By enhancing reliability beyond in-distribution scenarios, these results hold significant implications for the wider adoption of artificial intelligence in safety-critical applications.

\end{abstract}


\section{Introduction}
\label{introduction}

Over the past decade, deep learning has revolutionised pivotal industries, ranging from natural language processing \cite{vaswani2017attention,guo2025deepseek} and autonomous driving \cite{badue2021self,dhaif2024review} to drug discovery \cite{jumper2021highly,blanco2023role}. Despite vast data quantities required for training, unfamiliar testing scenarios undermining model reliability persist. For instance, a language model trained exclusively on English text may fail catastrophically on French queries, exemplifying the out-of-distribution (OoD) challenge. Within these problems, very few have captured the interest of the scientific community as much as cases where testing and training supports are disjoint, which we characterise as \emph{out-of-support} (OoS).

Traditional OoS solutions rely on incorporating inductive biases, such as enforcing known dynamics \cite{nzoyem2025neural,rackauckas2020universal} or prioritising discriminative features \cite{keshtmand2026counterfactual}. However, these methods falter when valid inductive biases are unavailable. Strategies like Distributionally Robust Optimisation \cite{kuhn2025distributionally} and meta-learning \cite{caruana1997multitask,hospedales2021meta} similarly necessitate prior knowledge of potential test distributions. While non-parametric approaches like Gaussian Processes \cite{williams1995gaussian} offer inherent uncertainty estimates, they scale poorly to large datasets.

Leveraging recent advances in sequence modelling and weight-space learning \cite{schurholt2024towards,nzoyem2025weight}, we propose \textbf{\themethod}: a framework recasting OoS generalisation tasks as forecasting problems through the partitioning of the training set into concentric \emph{shells} that we call ``rings''. By mapping each ring to a discrete time step, we learn sequential dynamics extrapolatable to the test set. Our contributions are threefold: (1) a computationally efficient, parametric, interpretable, and inductive bias-free framework for OoS generalisation; (2) a linearisation strategy enabling the framework to provide uncertainty estimates both in-distribution (InD) and OoS; and (3) empirical validation on synthetic sinusoidal and real-world air quality experiments revealing superior or competitive performance at low parameter count.


\subsection{Problem Setting}

In this typical machine learning problem, we are interested in learning the parametrised mapping $f_{\theta}: \mathbb{R}^{D_x} \mapsto \mathbb{R}^{D_y}$ such that $y = f_{\theta}(x),  \forall (x,y) \sim p$, where $\theta$ is the array of model parameters, and $p$ denotes the joint probability distribution of the data. We denote by $p_{\text{tr}}$ the training data distribution from which a \emph{fixed} number of training inputs $X^{\text{tr}} \in \mathbb{R}^{N_{\text{tr}} \times D_x}$ are sampled along with their corresponding outputs $Y^{\text{tr}} \in \mathcal{Y}$.\footnote{$\mathcal{Y}$ could be a subset of $\mathbb{R}^{D_y}$ for regression, or a discrete set such as $\{ 0, 1\}$ for classification.}

OoS requires models to maintain predictive power on input samples $X^{\text{te}} \in \mathbb{R}^{N_{\text{te}} \times D_x}$ from $p_{\text{te}}$ defined in regions of the input space where the training density is zero. Concretely, this means $\text{Supp} (X^{\text{tr}}) \cap  \text{Supp}(X^{\text{te}}) = \emptyset$, where $\text{Supp}$ denotes the support of a discrete set of points, i.e., the range of $\mathbb{R}^{D_x}$ covered by these points.

\subsection{Related Work}

OoS problems have traditionally been studied under the umbrella of OoD generalisation. Distributionally Robust Optimisation (DRO) \cite{kuhn2025distributionally} remains one of the strongest approaches, as it trains models to perform well under worst-case scenarios. DRO, however, requires the specification of a complex ambiguity set of possible testing distributions $\testdist$. Engression \cite{shen2025engression} is designed to bridge the gap between traditional regression and full distributional modelling, specifically targeting the support-shift limitations of standard DRO. Beyond simple point estimates, Engression captures the stochastic nature of the data generation process itself.

Recent literature has shifted toward Invariant Learning \cite{arjovsky2020out} and causal discovery to identify features that remain relevant outside the training support \cite{keshtmand2026counterfactual}. Concurrently, non-parametric methods, most notably Gaussian Processes (GPs) \cite{williams1995gaussian}, have been revitalised for OoS tasks due to their principled approach to uncertainty. Unlike most parametric models that collapse in unseen regions, non-parametric approaches revert to a prior belief when the test data escapes the support of the training set. This comes at a cost, and our approach addresses the main limitation of GPs by requiring significantly less computational resources to fit large training sets $X^{\text{tr}}$.

Another family of OoS generalisation techniques is meta-learning \cite{caruana1997multitask,hospedales2021meta,nzoyem2025learning,finn2017model}, which aims to fine-tune (a subset of) the model's parameters $\theta$ at test time on a new task. Like DRO, however, a successful meta-learner requires some inductive bias of what the target testing distribution could be. In contrast, our approach attempts to extrapolate beyond the training domain by providing the most likely OoS predictions under a sequence model assumption, thereby eliminating the need for explicit inductive biases and test-time fine-tuning.

Our approach is also heavily connected to the nascent area of weight-space learning (WSL) \cite{schurholt2024towards}. Remarking on the growing capacity of public model repositories such as CivitAI and HuggingFace \cite{jain2022hugging}, WSL has recently captured the attention of the deep learning community, showcasing huge breakthroughs in implicit neural representations \cite{dupont2022data}, generalisation error prediction \cite{unterthiner2020predicting}, and sequence modelling \cite{nzoyem2025weight}.

\section{Method}
\label{method}

\begin{figure}[h]
    \centering
    \includegraphics[width=\linewidth]{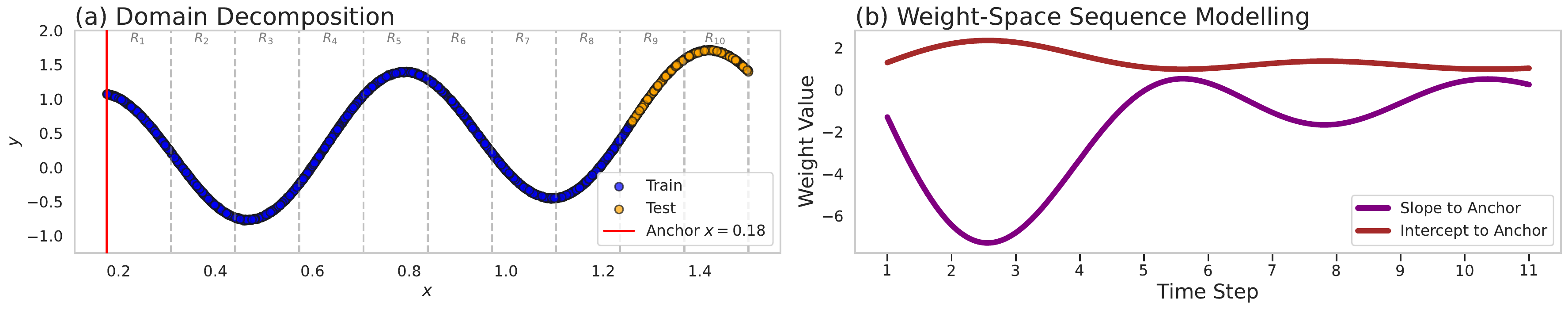}
    \caption{Illustration of the two main steps of the \themethod framework for sinusoidal extrapolation. \textbf{(a)} First, an anchor point is chosen and the input domain $\mathbb{R}^1$ is decomposed into $T=10$ ``rings'', here clearly delineated as intervals. Within each ring, we consider a simple linear model $\hat y = \theta^1 \cdot x + \theta^2$, where $\theta = [ \theta^1, \theta^2 ]^T$ contains the slope and intercept to anchor, respectively. \textbf{(b)} Optimal weights $\{\theta_t\}_{t=1}^{T_{\text{tr}}}$ for the data in each ring are subsequently computed by fitting a weight-space sequence model, one ring corresponding to one time step. Suitable weights for OoS datapoints are obtained by rolling out the sequence model for time steps $t \geq T_{\text{tr}}$ (in this example, $T_{\text{tr}}=9$).}
    \label{fig:osseq}
\end{figure}

\cref{fig:osseq} summarises our proposed \textbf{\themethod}using a 1D sine wave as our example. We begin by selecting an anchor point from the training dataset, and we decompose the training domain into successive \emph{hyperspherical shells} (equivalent to intervals in 1D or annuli in 2D). While their geometric structure might be different depending on $D_{x}$ and the chosen distance metric, we refer to these shells as ``rings'' for simplicity. Each ring corresponds to a step in a sequence model, and a weight-space sequence model learns to predict suitable weights for each step.


In the following paragraphs of this section, we detail the ingredients of the \themethod framework. We begin by outlining the input domain decomposition and weight-space sequence modelling strategies. Deterministic training and inference algorithms are discussed, followed by a stochastic framework for uncertainty estimation.

\subsection{Domain Decomposition}

Domain decomposition is the first stage of the \themethod framework. Since the input domain is a metric space, we can define a distance metric $d(\cdot, \cdot)$ on $\mathbb{R}^{D_x}$. Next, an anchor point $\underline{x}$ is chosen. For each point in either the training or testing dataset, we can compute its distance to $\underline{x}$.

Depending on the desired granularity, we construct $T>1$ rings $\{\textsfit{R}_t\}_{t=1}^T$ of \emph{equal} radii $\delta$. Using precomputed distances to the anchor, we assign a ring identifier to each data point. The data within the ring $\textsfit{R}_t$ is denoted $X_{t}^{\text{tr}}$ for train samples, and $X_{t}^{\text{te}}$ for testing samples. Necessarily, $T_\text{tr}$ that identifies the outermost ring containing a training datapoint is such that $T_\text{tr} \leq T$.

Conventional machine learning seeks to learn a single model $\hat y = f_{\theta}(x) $,  $\forall x \in X^{\text{tr}} \triangleq \bigcup_{t=1}^{T_{\text{tr}}} X_{t}^{\text{tr}}$. In contrast, \themethod seeks to fit one model $\theta_t$ for each ring, such that $\hat y = f_{\theta_t}(x), \forall x \in X_{t}^{\text{tr}} $. We achieve this via weight-space sequence modelling.

\subsection{Weight-Space Sequence Modelling}

The domain decomposition above naturally provides a sequential ordering for the weights $\theta_t$, which we use to formulate an initial value problem (IVP). Specifically, our optimisation procedure is the weight-space learning problem defined as follows\footnote{Note the use of the sum and not the average; this is because some rings might be empty.}
\begin{equation} \label{eq:recloss}
\phi^*, \theta_1^* = \argmin_{\phi, \theta_1} \sum_{t=1}^{T_{\text{tr}}} \mathop{\mathbb{E}}\limits_{\tiny(x, y) \sim (X_{t}^{\text{tr}}, Y_{t}^{\text{tr}})} \left[ \ell(f_{\theta_t}(x), y) \right], \quad \text{subject to} \quad \{\theta_t \}_{t=2}^{T_{\text{tr}}} = G_{\phi}(\theta_1),
\end{equation}
where:
\begin{itemize}
    \item $\ell(\cdot, \cdot)$ is a discrepancy function, such as mean squared error (MSE) or cross-entropy;
    \item $G_{\phi}(\cdot)$ is a higher level neural functional, a \emph{state-to-sequence}\footnote{Conventional \emph{sequence-to-sequence} models such as Transformers, SSMs, LSTMs, WSL-RNN are equally suitable for this task \cite{nzoyem2025learning}.} model parametrised by $\phi$.
\end{itemize}
Our goal is to not only find optimal parameters $\phi^*$, but also the initial weights $\theta_1^*$ for the IVP via gradient descent. Note how \cref{eq:recloss} provides no supervision signal beyond $T_{\text{tr}}$. This is at the core of our approach, as the sequence model $G_{\phi}$ will predict OoS weights $\{ \theta_t \}_{t=T_{\text{tr}}+1}^T$ using the same dynamics that were learned within the training domain. 


By learning the dynamics of the weights $\theta_t$ through exposure to disjoint but consecutive rings, \themethod effectively takes a \emph{global} view of the data and learns a generalisable predictive distribution, rather than the \emph{local} view taken by conventional deep learning. The overall algorithms for training and testing are presented in \cref{alg:oosseq}.

\begin{algorithm}[h]
\caption{\themethod Training and Inference}
\label{alg:oosseq}

\begin{minipage}{0.52\textwidth}
\centering
\vspace*{-1mm}
\textbf{Training}
\vspace{1mm}
\begin{algorithmic}[1]
\State \textbf{Input:} Data $X^{\text{tr}}, Y^{\text{tr}}$,  anchor $\underline{x}$, distance $d (\cdot, \cdot)$, \\ \hspace*{1.1cm}ring width $\delta$, random $\phi$ and $\theta_1$, \\ \hspace*{1.1cm}batch size $B$ 
\State \textbf{Output:} Trained $\phi$, $\theta_1$

\State \phantom{blank line for alignment}
\State \hspace{0.1cm} Partition $X^{\text{tr}}$ into subsets $\{X^{\text{tr}}_t\}_{t=1}^{T_{\text{tr}}}$ using \\ \hspace{0.1cm} $d(\underline{x}, \cdot)$ and $\delta$

\State \phantom{blank line for alignment}
\State \hspace{0.1cm} $\{\theta\}_{t=2}^{T} \gets G_{\phi}( \theta_1 )$

\State \phantom{blank line for alignment}
\While{not converged}
    \State $\mathcal{L}_{\text{tot}} \gets 0$
    \For{$t = 1$ \textbf{to} $T_{\text{tr}}$}
        \State $(X, Y) \gets \text{Subsample}(X_t^{\text{tr}}, Y_t^{\text{tr}}, B)$
        \State $\mathcal{L}_{\text{tot}} \gets \mathcal{L}_{\text{tot}} + \frac{1}{B} \sum\limits_{x,y \in X, Y} \ell(f_{\theta_t}(x), y)$
    \EndFor
    \State Update $\phi, \theta_1$ via $\nabla \mathcal{L}_{\text{total}}$
\EndWhile
\end{algorithmic}
\end{minipage}
\hfill
\vrule
\hfill
\begin{minipage}{0.46\textwidth}
\centering
\vspace{2mm}
\textbf{Inference}
\vspace{1mm}
\begin{algorithmic}[1]
\State \textbf{Input:} Test point $x$, anchor $\underline{x}$, \\ \hspace{1.1cm}distance $d(\cdot, \cdot)$,  ring width $\delta$, \\ \hspace{1.1cm}trained $\phi$ and $\theta_1$
\State \textbf{Output:} Prediction $\hat y$

\State \phantom{blank line for alignment}
\State \hspace{0.1cm} $d_{\text{test}} = d(\underline{x}, x)$

\State \phantom{blank line for alignment}
\State \hspace{0.1cm} $t_{\text{test}} = \lfloor d_{\text{test}} / \delta \rfloor$

\State \phantom{blank line for alignment}

\State \hspace{0.1cm} $\{\theta\}_{t=2}^{t_\text{test}} \gets G_{\phi}( \theta_1 )$

\State  \phantom{blank line for alignment}
\State \hspace{0.1cm} $\hat{y} = f_{\theta_{t_{\text{test}}}}(x)$

\end{algorithmic}
\vspace{2.7cm}
\end{minipage}

\end{algorithm}

\subsection{Stochastic Framework for Regression}

To handle uncertainty, we extend the weight-space sequence model to a stochastic framework. Instead of a point estimate $\theta_t$, the $G_{\phi}$ outputs the parameters of a distribution over weights. This approach allows us to propagate uncertainty from the weight space to the prediction space.

\paragraph{Reparameterisation trick.} We assume the weights at time step $t$ follow a Gaussian distribution $\theta_t \sim q(\theta_t) = \mathcal{N}({\mu}_t, \text{diag}({\sigma}^2_t))$. To enable backpropagation through the sampling process, we employ the reparameterisation trick \cite{kingma2013auto}
\begin{equation}\theta_t = {\mu}_t + {\sigma}_t \odot {\epsilon}, \quad {\epsilon} \sim \mathcal{N}(0, I)\end{equation}
where ${\mu}_t$ and ${\sigma}_t$ are the mean and standard deviation predicted by $G_{\phi}$, and $\odot$ denotes the element-wise product. $0$ and $I$ are respectively the zero vector and the intentiy matrix of suitable dimensions.

\paragraph{Marginalisation via linearisation.} A primary challenge in OoS generalization is obtaining the predictive distribution $p(y | x)$, which requires marginalising over the weights
\begin{equation}p(y | x) = \int p(y | x, \theta) q(\theta) \md \theta.\end{equation}
Since this integral is analytically intractable for deep neural networks, we perform a first-order Taylor expansion of the model $f_{\theta}(x)$ around the mean weights ${\mu_t}$
\begin{equation}f_{\theta}(x) \approx f_{{\mu}_t}(x) + \mathbf{J} (\theta - {\mu}_t),\end{equation}
where $\mathbf{J} = \left. \frac{\partial f_{\theta}(x)}{\partial \theta} \right|_{\theta = \bm{\mu}_t}$ is the Jacobian of the model outputs with respect to the weights. Under this linear approximation, we write the predictive distribution as
\begin{equation}\hat{y} \sim \mathcal{N}(\mu_y, \Sigma_y) \quad \text{with} \quad\begin{cases}\mu_y = f_{{\mu}_t}(x),  \\
\Sigma_y = \mathbf{J} \text{diag}({\sigma}^2_t) \mathbf{J}^\top + \sigma_{\text{noise}}^2 I,\end{cases}\end{equation}
where $\sigma_{\text{noise}}$ is a hyperparameter included for numerical stability and to account for underlying noise in the ground truth output measurements. This allows us to obtain not only mean predictions but also a principled covariance $\Sigma_y$ that reflects model uncertainty.

\paragraph{Loss function regularisation.} To prevent the model from producing overconfident predictions in OoS regions, we regularise the loss function in \cref{eq:recloss}. We introduce a KL divergence term between the predicted distribution and a standard Gaussian prior $p(\hat{y}) = \mathcal{N}(0, I)$. The loss function becomes
\begin{equation}\mathcal{L}(\phi, \mu_1, \sigma_1) = \sum_{t=1}^{T_{\text{tr}}} \mathop{\mathbb{E}}\limits_{\tiny(x, y) \sim (X_{t}^{\text{tr}}, Y_{t}^{\text{tr}})} \left[ \ell(f_{\mu_t}(x), y) \right] + \beta \cdot D_{\text{KL}} \left( \mathcal{N}(\mu_y, \Sigma_y) | \mathcal{N}(0, I) \right),\end{equation}
where $\mu_1, \sigma_1$ are respectively the learned initial mean and standard deviation weights, and $\beta$ is a scaling hyperparameter. This loss promotes a graceful reversal toward the prior rather than collapsing as the model moves further from the training support.

\section{Main Results}
\label{experiments}

\subsection{Experimental Setup}

We evaluate \themethod on two regression benchmarks designed for OoS generalisation: a synthetic periodic function and a real-world sensor correlation task. 

\paragraph{Cosine Dataset.} A classic 1D regression problem where $y = \cos(10x) + 0.5x + \epsilon$, where $\epsilon \sim \mathcal{N}(0,25e-6)$. The data is partitioned into disjoint train and test sets. This task requires the model to extrapolate the trend and periodicity into the unseen intervals. We set $\beta=1e-2$. The anchor $\underline{x}$ is chosen as the \emph{average} across $X^{\text{tr}}$. We specify $T=600$ rings to simulate the weight-space sequence, of which $T_{\text{tr}}=300$ overlap the training data. 


\paragraph{AirQuality Dataset.} Derived from the UCI Air Quality dataset \cite{air_quality_360}, we model the relationship between two chemical sensors: PT08.S5 (O3) as the input $x$ and PT08.S3 (NOx) as the target $y$. Once normalised, we split the data based on a threshold of the $O_3$ sensor readings ($x > 1$), creating a distinct support shift between training and testing distributions. We use $\beta=5e-2$. The anchor $\underline{x}$ is chosen as the \emph{minimum} across $X^{\text{tr}}$. For this dataset, we set $T=80$ and $T_{\text{tr}}=40$.

For both regression tasks, we use a linear regression model $f_{\theta}(x) = x \cdot \theta^1 + \theta^2$ with 2 scalar learnable parameters. For the sequence model $G_{\phi}$, we considered an autoregressive linear recurrence in weights space $\theta_{t+1} = \phi \theta_t$, with $\phi \in \mathbb{R}^{2\times 2}$ as a learnable matrix. The discrepancy metric $\ell(\cdot, \cdot)$ is the MSE. Training \themethod is performed using the Adabelief optimiser \cite{zhuang2020adabelief} within the JAX ecosystem \cite{jax2018github}. We consider three baselines: a standard MLP \cite{mcculloch1943logical}, a Gaussian Process \cite{williams1995gaussian}, and an Engression model \cite{shen2025engression}.


\begin{figure}[ht]
    \centering
    \includegraphics[width=0.245\linewidth]{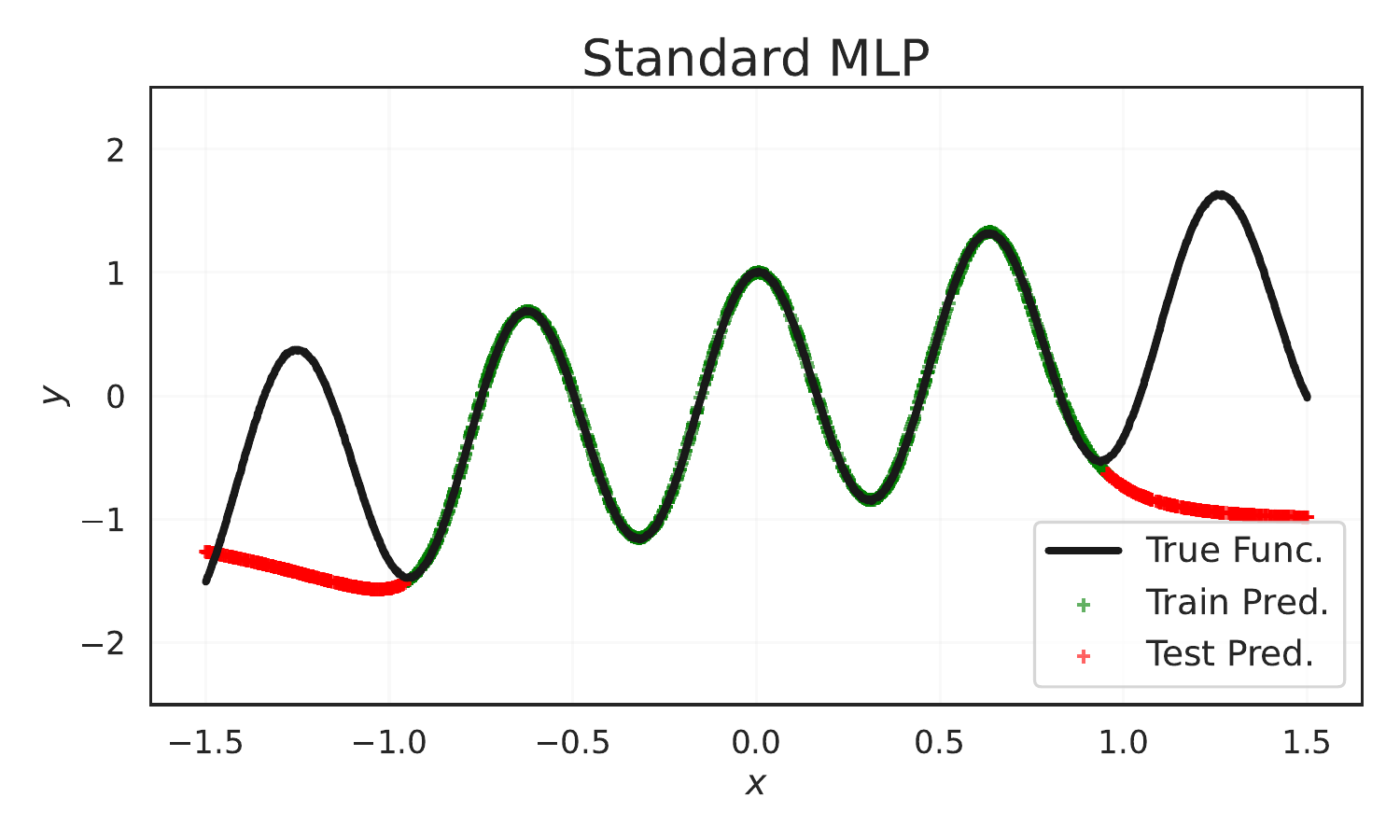}
    \includegraphics[width=0.245\linewidth]{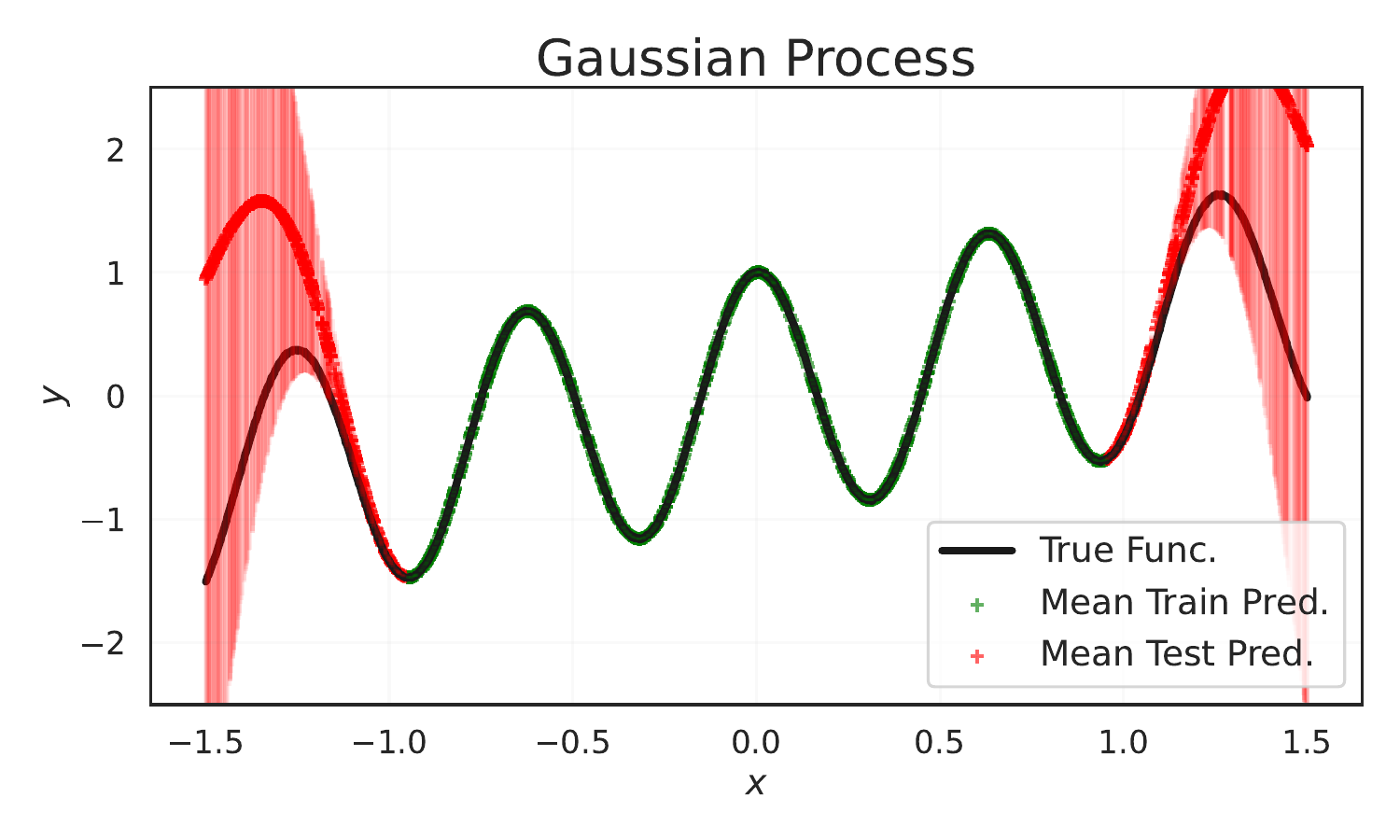}
    \includegraphics[width=0.245\linewidth]{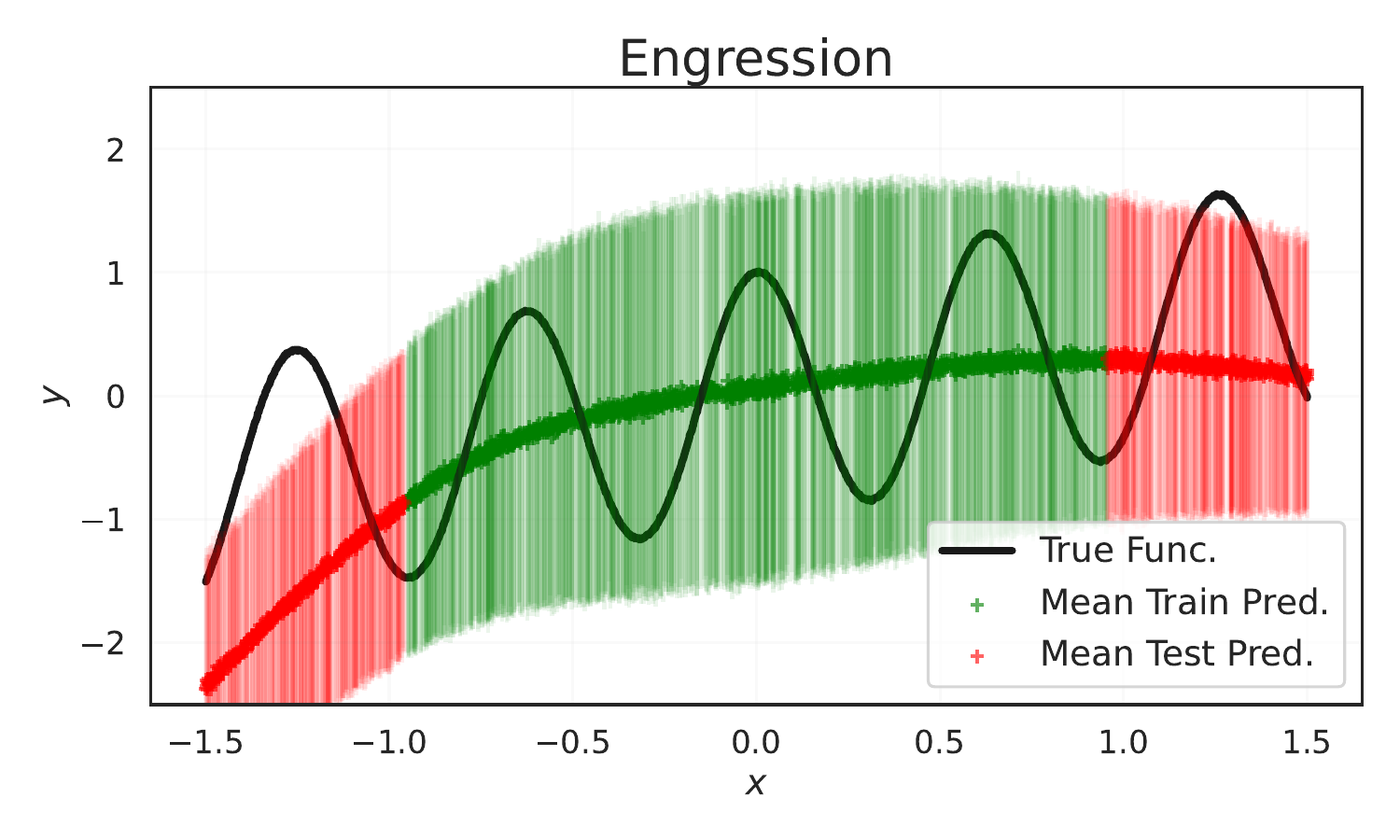}
    \includegraphics[width=0.245\linewidth]{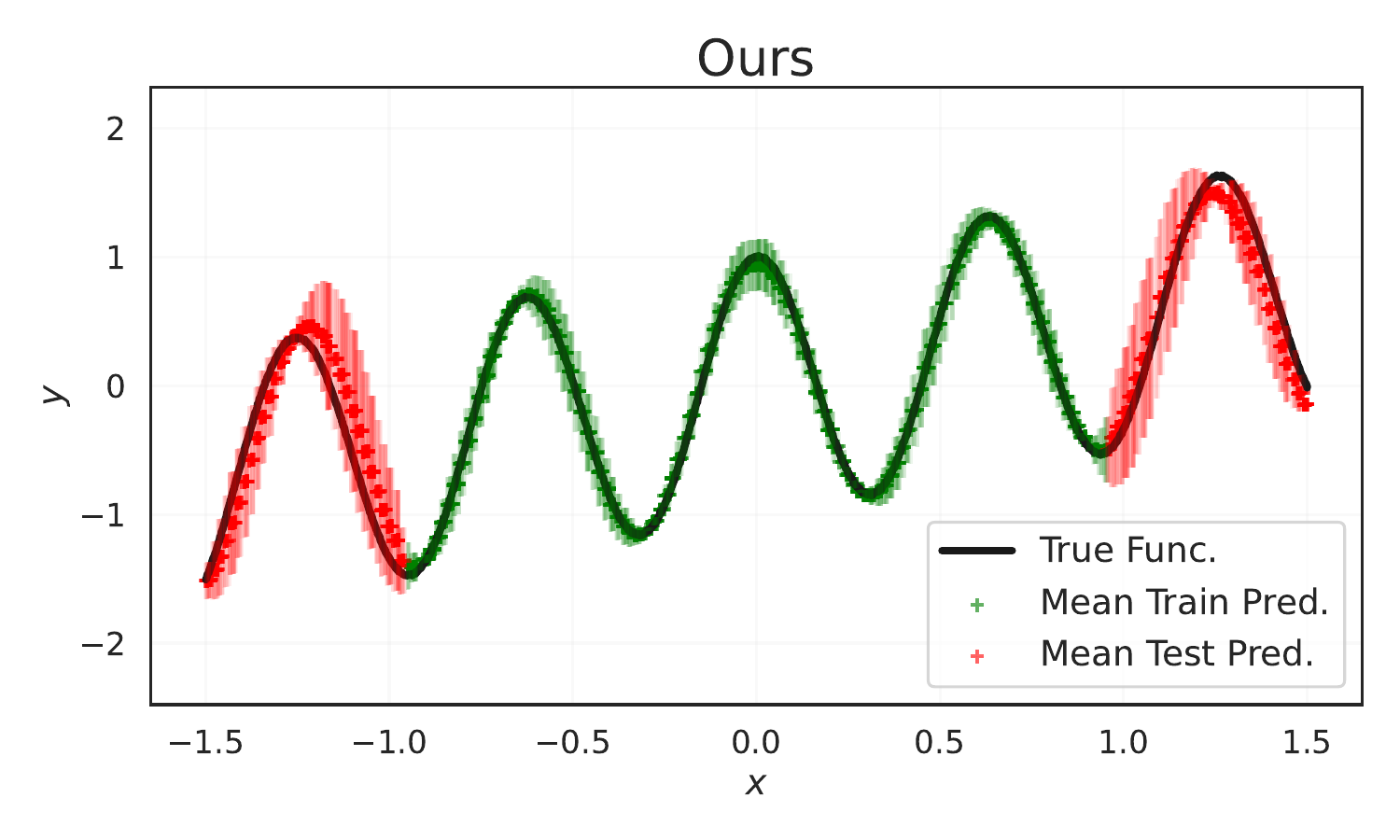}
    \includegraphics[width=0.245\linewidth]{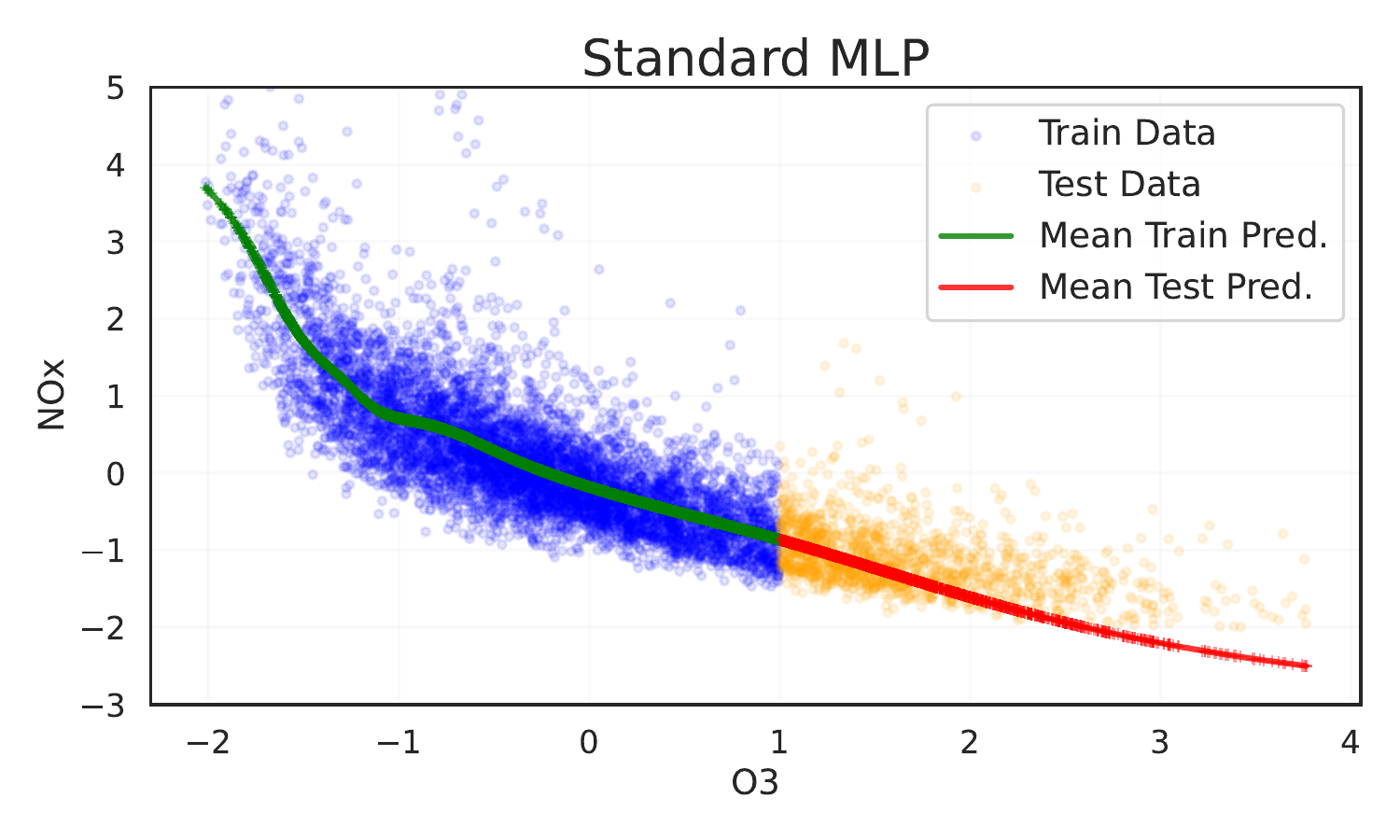}
    \includegraphics[width=0.245\linewidth]{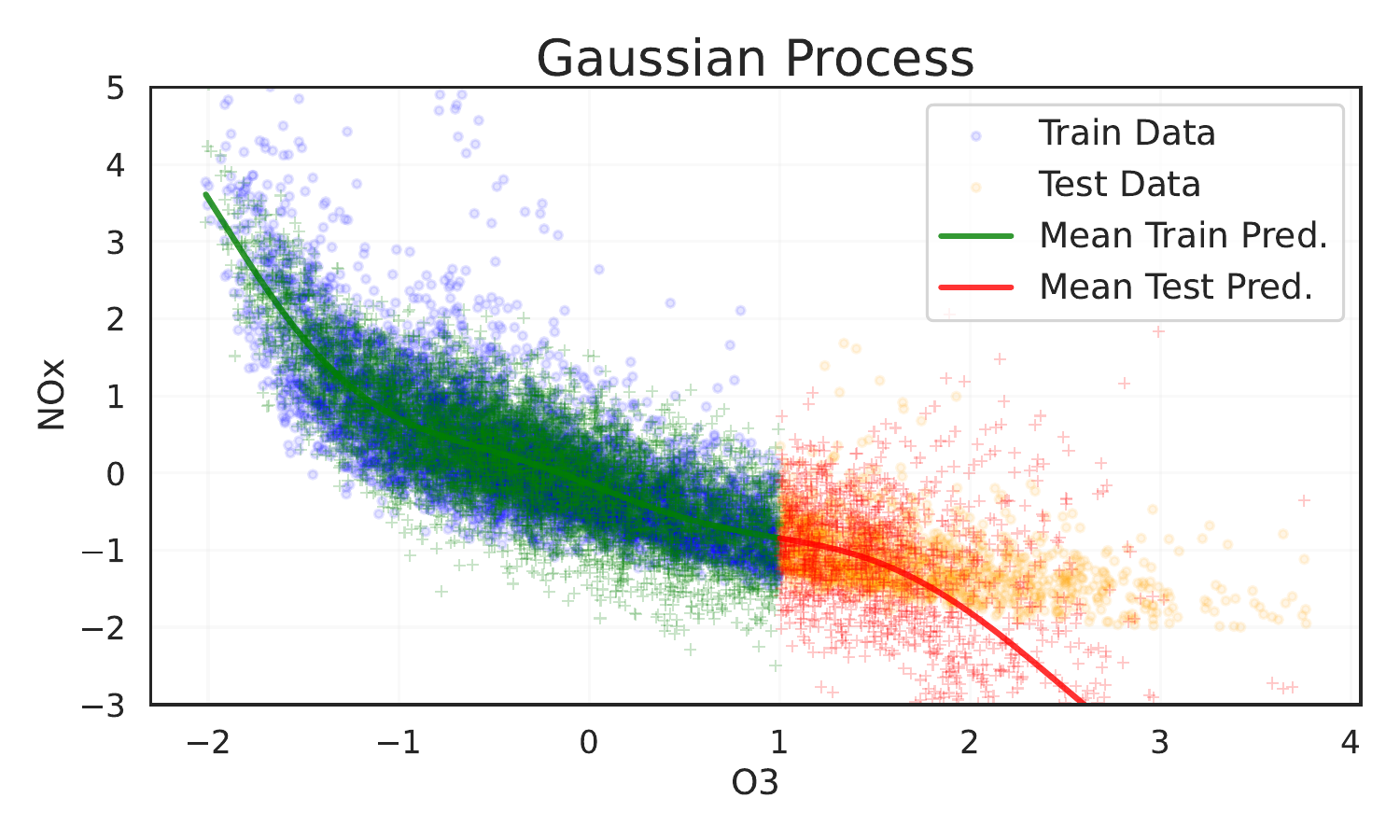}
    \includegraphics[width=0.245\linewidth]{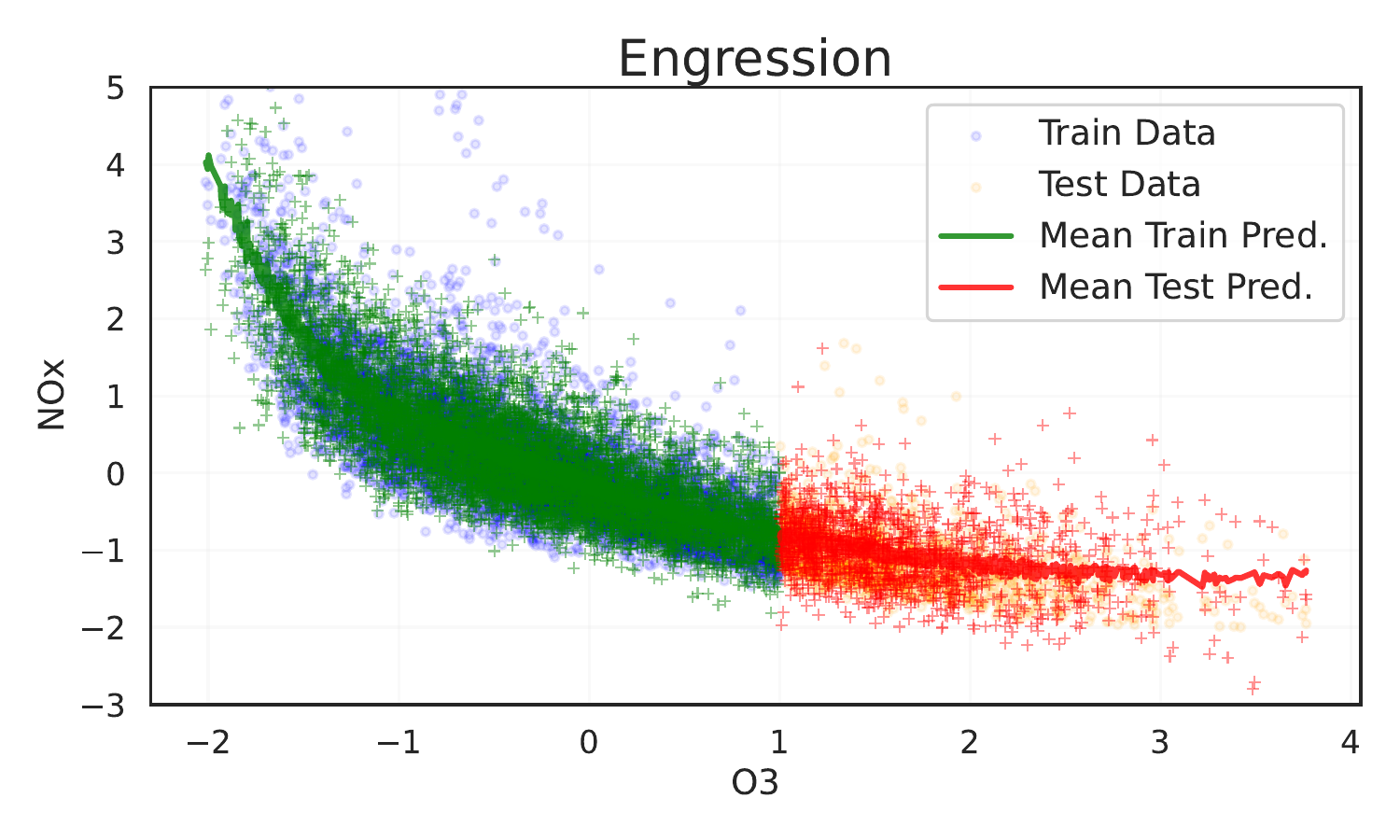}
    \includegraphics[width=0.245\linewidth]{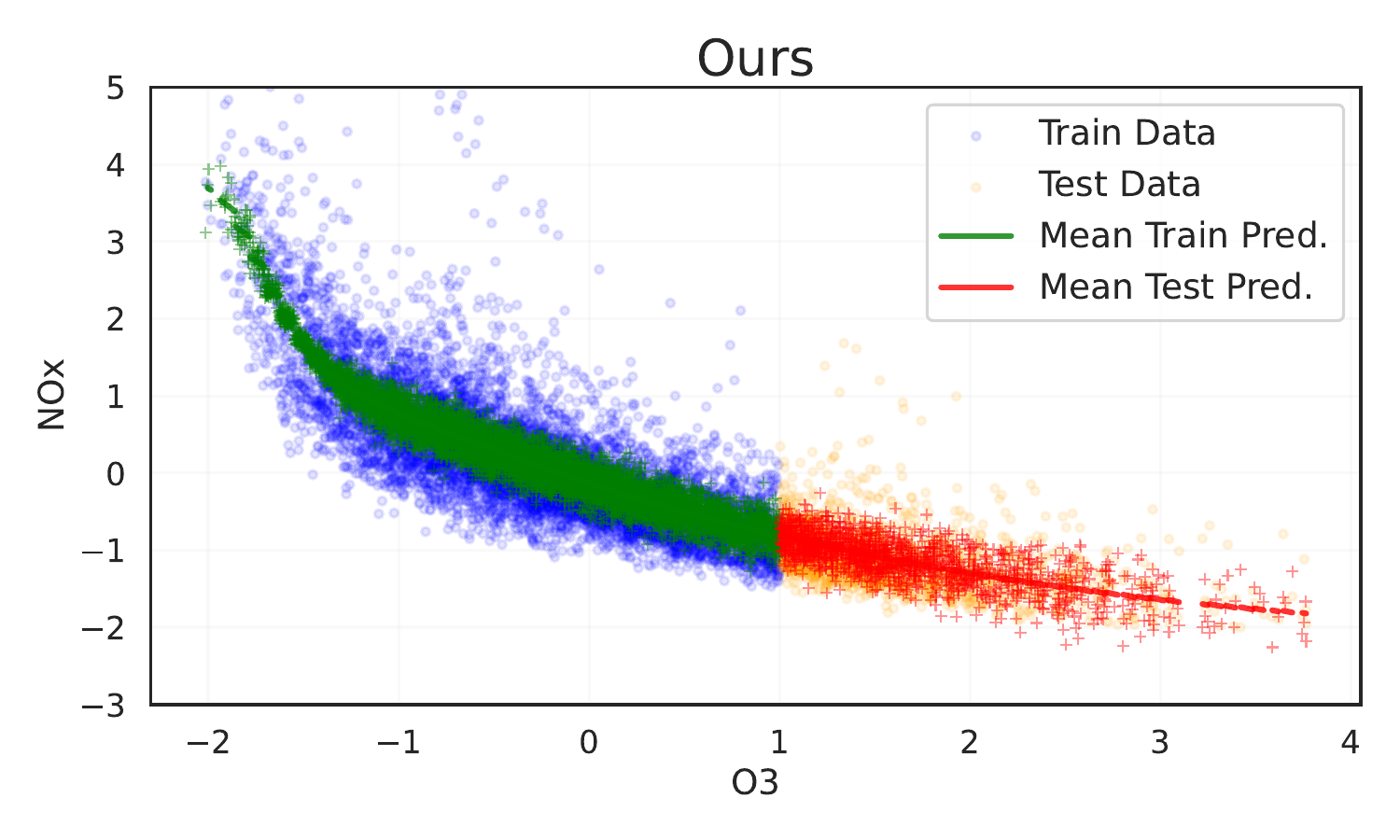}
    \caption{Performance of \themethod compared to baselines. (Top Row) Extrapolation on the Cosine wave experiment. (Bottom Row) OoS generalisation on the AirQuality sensor dataset. Shaded areas represent the pointwise $2\sigma$ uncertainty estimates.}
    \label{fig:results}
\end{figure}

\begin{table}[ht]
    \centering
    \small
    \caption{Train (in-distribution) and test (out-of-support) MSEs across datasets ($\downarrow$).}
    \label{tab:results}
    \begin{tabular}{l c c c c}
        \toprule
        \textbf{Method} & \textbf{Cosine (InD)} & \textbf{Cosine (OoS)} & \textbf{AirQuality (InD)} & \textbf{AirQuality (OoS)} \\
        \midrule
        Standard MLP & 0.00021 & 2.3672 & 0.3175 & 0.2284 \\
        Gaussian Process & 0.00002 & 1.3973 & 0.3203 & 0.7053 \\
        Engression & 0.50802 & 1.3240 & 0.3240 & 0.1603\\
        \textbf{\themethod (Ours)} & 0.00190 & 0.3502 & 0.3484 & 0.1381 \\
        \bottomrule
    \end{tabular}
\end{table}

\subsection{Discussion}
\label{discussion}

The results in \cref{fig:results,tab:results} demonstrate that \themethod effectively sidesteps the catastrophic extrapolation failures common in standard neural networks. By reformulating the problem as weight-space sequence modelling, we allow the model to learn the evolution of the function's parameters.

The success in the \textbf{Cosine} experiment suggests that our sequence model captures the underlying periodicity of the weight trajectory, allowing it to ``forecast'' the weights of the outermost rings accurately. Interestingly, Engression fails to capture the conditional distribution $p(y|x)$, neither in-distribution nor out-of-distribution.

In the \textbf{AirQuality} task, \themethod performs on-par with state-of-art techniques like Engression, especially on the testing OoS data. Unlike non-parametric methods like Gaussian Processes that scale poorly with the training data size, \themethod maintains the computational efficiency of parametric models while providing competitive uncertainty estimates via linearisation.


Crucially, \themethod exhibits such powerful performance at extremely low parameter count $D_{\themethod} = D_{\theta} + D_{\phi} = D_{\theta} + D_{\theta}^2 = 2+4=6 $.\footnote{In our recurrent implementation, we use an ANODE-style augmentation \cite{dupont2019augmented} for increased representational power, resulting in slightly larger parameters $D_{\phi} = (D_{\theta} + a)^2$, with $a$ the augmentation size.} In effect, \themethod seeks to fit each $\theta_t$ on a fraction of the original dataset, which requires less parameters, thus limiting the size of the regression model $\theta$. This translates to strong computational and memory savings, underscoring a significant benefit absent in other parametric models such as the standard MLP or Engression.

Additionally, the square matrix $\phi$ in the linear recurrence $G_{\phi}$ captures the weight dynamics. An eigendecomposition of this matrix would reveal important characteristics for generalisation to a (theoretical) infinitely wide out-of-support domain. This makes our framework highly interpretable, which is critical when deploying AI models in the real-world.

Conversely, the error bars in \cref{fig:results} suggest that \themethod tends to be just as confident in the training domain as it is out-of-support. While this can be addressed by careful tuning of the scaling hyperparameter $\beta$, it displays a critical limitation in that \themethod introduces several such hyperparameters (e.g., $d, \delta, \underline{x}, \beta$ as seen in \cref{alg:oosseq}), all of which require tuning to achieve satisfying results.

\section{Conclusion}
\label{conclusion}

In this paper, we introduced \themethod, a framework that transforms out-of-support generalisation into a weight-space sequence forecasting task. By decomposing the input space into concentric rings and modelling the trajectory of optimal weights, we enable reliable extrapolation with no inductive biases. Our stochastic formulation further provides uncertainty estimates through model linearisation and output-space KL regularization. Experimental results on periodic synthetic data and real-world air quality sensors confirm that \themethod outperforms standard regression baselines and competitive generalisation methods. Future work will explore the theoretical underpinnings of this approach in the infinite-length $T\rightarrow \infty$ regime, along with the scaling to high-dimensional manifold data. Chosing an appropriate anchor location $\underline{x}$ remains a challenge we wish to solve. We will also explore proven strategies for more confident in-distribution uncertainties while simultaneously requiring conservative out-of-support estimates.



\subsection*{Broader Impact}
\label{sec:impact}
By enabling generalisation to unseen scenarios, this work could contribute to the mitigation of catastrophic failures in critical sectors such as environmental monitoring, healthcare, and infrastructure management. Our approach encourages a more transparent understanding of a model's operational limits, fostering the responsible deployment of predictive AI. To accelerate usability and foster reproducibility, we provide our code at \url{https://github.com/ddrous/weightcaster}.

\subsubsection*{Acknowledgments} 
This work was supported by UK Research and Innovation grant EP/S022937/1: Interactive Artificial Intelligence.



\bibliography{refs}
\bibliographystyle{plainnat}

\newpage


\appendix


\end{document}